\pdfoutput=1

\documentclass[11pt]{article}

\usepackage[final]{acl}

\usepackage{times}
\usepackage{latexsym}

\usepackage[T1]{fontenc}

\usepackage[utf8]{inputenc}

\usepackage{microtype}

\usepackage{inconsolata}

\usepackage{graphicx}

\usepackage{hyperref}       
\usepackage{url}            
\usepackage{booktabs}       
\usepackage{amsfonts}       
\usepackage{nicefrac}       
\usepackage{microtype}      
\usepackage{lipsum}		
\usepackage{natbib}
\usepackage{doi}
\usepackage{algorithm}
\usepackage{algorithmic}
\usepackage{amssymb}
\usepackage{xcolor}         
\usepackage{subfigure}
\usepackage{amsmath}
\usepackage{amsthm}
\usepackage{multirow}
\usepackage{diagbox}
\usepackage{enumitem}
\usepackage{bm}
\usepackage{color, colortbl}
\usepackage{subcaption}

\newcommand{\NAME}{\textbf{ALPS} }
\definecolor{COLOR_MEAN}{HTML}{f0f0f0}

\title{ALPS: Attention Localization and Pruning Strategy for Efficient Alignment of Large Language Models}

\author{Hao Chen\textsuperscript{$\spadesuit$}, \ Haoze Li\textsuperscript{$\spadesuit$}, \ Zhiqing Xiao\textsuperscript{$\spadesuit$}, \ Lirong Gao\textsuperscript{$\spadesuit$}, \ Qi Zhang\textsuperscript{$\spadesuit$}\\
\ \textbf{Xiaomeng Hu\textsuperscript{$\spadesuit$}}, \ \textbf{Ningtao Wang\textsuperscript{$\clubsuit$}}, \ \textbf{Xing Fu\textsuperscript{$\clubsuit$}}, \ \textbf{Junbo Zhao\textsuperscript{$\spadesuit$}} \\
  \textsuperscript{$\spadesuit$}Zhejiang University \quad
  \textsuperscript{$\clubsuit$}Ant Group \\
  \texttt{\{h.c.chen, j.zhao\}@zju.edu.cn}
  } 

\begin{document}
\maketitle
\begin{abstract}
Aligning general-purpose large language models (LLMs) to downstream tasks often incurs significant training adjustment costs.
Prior research has explored various avenues to enhance alignment efficiency, primarily through minimal-data training or data-driven activations to identify key attention heads.
However, these approaches inherently introduce data dependency, which hinders generalization and reusability.
To address this issue and enhance model alignment efficiency, we propose the \textit{\textbf{A}ttention \textbf{L}ocalization and \textbf{P}runing \textbf{S}trategy (\textbf{ALPS})}, an efficient algorithm that localizes the most task-sensitive attention heads and prunes by restricting attention training updates to these heads, thereby reducing alignment costs. 
Experimental results demonstrate that our method activates only \textbf{10\%} of attention parameters during fine-tuning while achieving a \textbf{2\%} performance improvement over baselines on three tasks.
Moreover, the identified task-specific heads are transferable across datasets and mitigate knowledge forgetting. 
Our work and findings provide a novel perspective on efficient LLM alignment.
The code is available at \href{https://github.com/VoiceBeer/ALPS}{https://github.com/VoiceBeer/ALPS}.

\end{abstract}

\section{Introduction}

\begin{figure}[t]
  \includegraphics[width=\columnwidth]{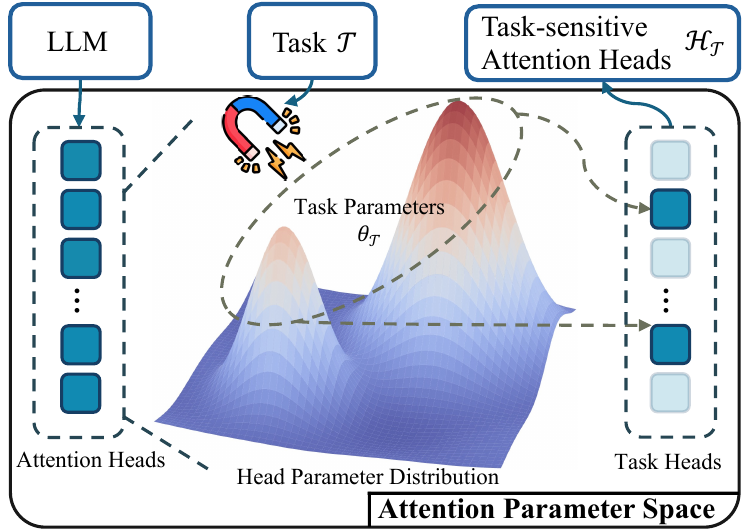}
  \caption{Process of localizing task-sensitive attention heads. The left side represents the attention heads of LLM, and the center illustrates the distribution of head parameters, where task parameters $\theta_{\mathcal{T}}$ capture task information and lead to task-sensitive heads $\mathcal{H}_{\mathcal{T}}$.}
  \label{fig:fig1}
\end{figure}

Pre-trained Large Language Models (LLMs) have demonstrated impressive performance across various tasks~\citep{achiam2023gpt,openai2025o3-mini,guo2025deepseek}.
A way to harness the potential of pre-trained foundation models on downstream tasks is task alignment, which mainly incorporates tailored task knowledge.
Recently, task alignment has garnered considerable attention in both industry and academia with
a diverse range of specialized models having been developed, spanning domains such as mathematics~\citep{shao2024deepseekmath,yang2024qwen2math}, code generation~\citep{guo2024deepseekcoder,roziere2023codellama}, medical diagnostics~\cite{singhal2023large,singhal2025toward} and protein structure prediction~\citep{schmirler2024fine}. 
However, despite the effectiveness of task alignment, it still poses a significant resource-intensive challenge.
The alignment process demands extensive efforts to construct task-specific instruction datasets and training adjustments, and further, its computational demands exacerbate the costs~\cite{zhao2023survey}.
These challenges highlight the pressing need for more efficient alignment strategies that not only minimize resource consumption costs but also maintain or even enhance model performance~\cite{wan2023efficient}.

\begin{figure*}[t]
  \includegraphics[width=\linewidth]{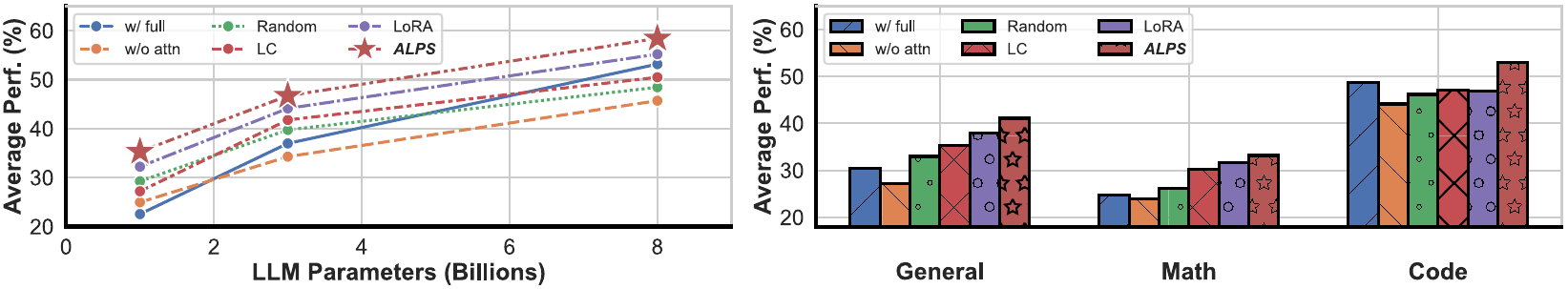}
  \caption{Preliminary results comparing \NAME against baselines across different LLM scales (\textbf{Left}), and downstream tasks (\textbf{Right}). \NAME consistently outperforms other methods, demonstrating its efficiency in diverse settings.}
  \label{fig:fig2}
\end{figure*}

To address the efficiency challenges of model alignment, 
recent LLM studies are interested in attention mechanisms and 
attempt to localize task-specific attention heads~\citep{clark2019does,michel2019sixteen,zhou2024role,wu2024retrieval,tang2024razorattention}.
However, these methods mainly rely on task-specific data to activate model parameters for attention head localization, inadvertently introducing data dependency to identified heads, which not only hinders generalization but also complicates reusability. 
Moreover, these approaches overlook the intrinsic functionality of the model weight parameters, which inherently encode task-relevant information~\citep{zhong2022theory,tam2024merging}, and can be leveraged to identify task-sensitive attention heads without relying on activation data.

In this work, to enhance alignment efficiency, utilize weight parameters, and introduce reusability of identified attention heads, we propose the \textit{\textbf{A}ttention \textbf{L}ocalization and \textbf{P}runing \textbf{S}trategy (\textbf{ALPS})}, a heuristic-guided algorithm that localizes task-sensitive attention heads and prunes by restricting attention training updates to these heads only. 
This algorithm draws inspiration from prior work leveraging model weight parameters to localize task-sensitive attention heads~\citep{voita2019analyzing,zheng2024attention,shi2024understanding}, as shown in Fig~\ref{fig:fig1}.
Unlike previous work utilizing whole model parameters, ALPS focuses on the distribution of each attention head. 
Specifically, given a base model and its task-related model, we extract and normalize the distribution of each head, then quantify the distributional shift between corresponding heads using our proposed parameter alignment distribution score. 
The heads with the highest scores are then selected as task-sensitive heads, while the remaining heads are frozen during fine-tuning.
This yields a pruned, task-specialized model with enhanced efficiency. 
Empirical results demonstrate that our method updates only \textbf{10\%} of attention parameters during fine-tuning while achieving a \textbf{2\%} performance gain over baselines across general, math, and code tasks, as shown in Figure~\ref{fig:fig2}. 
Further analysis reveals that the selected task-sensitive heads exhibit transferability and reusability, enhancing performance across datasets within the same task, reducing computational overhead, and avoiding data dependency. 
Moreover, the induced sparsity in attention heads helps mitigate knowledge forgetting in the model alignment.

In summary, our contributions are as follows:
\begin{itemize}
    \item We offer a novel perspective on efficient LLM alignment, pioneering an approach using model weight parameters to identify task-sensitive attention heads.
    \item We propose \textbf{ALPS}, a novel algorithm that identifies task-sensitive attention heads by measuring shifts in their weight distributions and then restricts training to these heads.
    \item Empirical results demonstrate that our method improves alignment efficiency and task performance while introducing sparsity and head transferability.
\end{itemize}

\section{Related Work}
\subsection{Efficient LLMs}
To mitigate the high costs of aligning foundation LLMs to downstream tasks, researchers have explored various strategies for efficient alignment~\citep{wan2023efficient}. For example, DeepSeek-R1~\citep{guo2025deepseek} relies solely on reinforcement training to reduce overhead. While data-focused approaches either aim to minimize the need for task-specific data~\citep{zhou2024lima,chen2023maybe,chen2023alpagasus} or augment downstream datasets~\citep{li2023self,wang2022self,xu2023wizardlm}, parameter-focused approaches improve efficiency by avoiding full-model updates~\citep{hu2021lora}, compressing the model~\citep{jacob2018quantization,hinton2015distilling}, and streamlining key-value storage~\citep{ainslie2023gqa,liu2024deepseek}. Our work leverages the attention mechanism to localize and prune task-sensitive attention heads.

\subsection{Attention Localization and Pruning}

\begin{figure*}[th]
  \includegraphics[width=\linewidth]{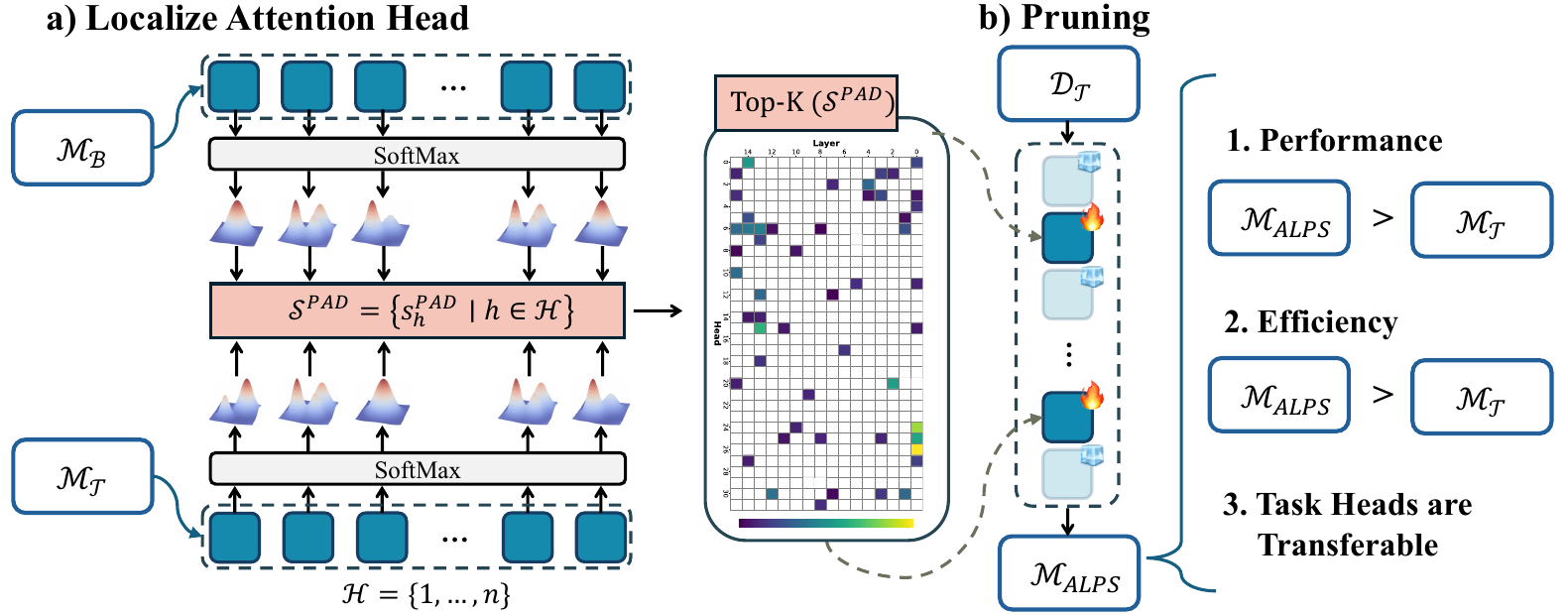}
  \caption {Overview of \NAME framework. \textbf{a)} Given a base model $\mathcal{M_B}$ and a task fine-tuned model $\mathcal{M_T}$, we extract their attention heads, compute weight distributions using softmax, and calculate their \textbf{PAD} score $s_{h}^{PAD}$. The Top-K attention heads are then selected based on $\mathcal{S}^{PAD}$, while \textbf{b)} pruning the remaining heads by freezing their gradient updates during fine-tuning. The resulting model $\mathcal{M}_{ALPS}$ outperforms $\mathcal{M_T}$ in both performance and efficiency, with transferable task-relevant heads that enhance alignment across datasets within the same task.}
  \label{fig:framework}
\end{figure*}

Recent studies have investigated attention mechanisms at various granularities to improve the efficiency and interpretability of LLM, including neurons, layers, and heads~\citep{geiger2021causal,gurnee2023finding,zou2023representation,zheng2024attention,zhao2024explainability}. Notably, many works have concentrated on attention heads for downstream tasks, including general linguistic and factuality capability~\citep{clark2019does,wu2024retrieval,zheng2024attention}, model safety~\citep{chen2024finding,zhou2024role}, KV-cache compression~\citep{ainslie2023gqa,fu2024not}, etc. Among them, ~\citet{michel2019sixteen} demonstrated that most attention heads can be dropped without compromising performance. While most methods rely on the activation of model parameters by feeding task-specific data, \citet{voita2019analyzing} analyzed the weight matrices of attention in models, revealing that specialized heads contribute to tasks, while the rest can be pruned. Similarly, \citet{shi2024understanding} and \citet{he2024matters} also observed that pruning certain attention layers enhances alignment efficiency, indicating redundancy within the attention mechanism. In contrast, our approach directly localizes task-relevant attention heads from weight matrices~\citep{voita2019analyzing,shi2024understanding}, thereby reducing data dependency and improving generalizability and efficiency in downstream alignment.

\section{Methodology} 
In this section, we first provide a preliminary overview of the attention mechanism with task parameters and then present \NAME in detail. Fig~\ref{fig:framework} illustrates the overall framework.

\subsection{Preliminary}
In this work, we adopt the Llama-3 series of models, which employ grouped-query attention (GQA) \citep{ainslie2023gqa} to reduce KV cache overhead. 
Formally, given an input sequence matrix $\bm{X} \in \mathbb{R}^{t \times d}$, where $t$ represents the sequence length, GQA divides the $n$ attention heads into $g$ groups, making all heads within the same group share a single pair of KV projections. Each head $h \in \{1,\dots,n\}$ learns projections 
\(\{\bm{W}_q^{h}, \bm{W}_k^{\lceil hg/n\rceil}, \bm{W}_v^{\lceil hg/n\rceil}\}\), where $\lceil \cdot \rceil$ is the ceiling function, and mapping $\bm{X}$ into queries \(\bm{Q}^{h}\), keys \(\bm{K}^{\lceil hg/n\rceil}\), and values \(\bm{V}^{\lceil hg/n\rceil}\):
\begin{equation}\label{eq:mha_qkv}
\begin{aligned}
    \bm{Q}^{h} &= \bm{XW}_q^{h} \in \mathbb{R}^{t \times d_k}, \\
    \bm{K}^{\lceil hg/n\rceil} &= \bm{XW}_k^{\lceil hg/n\rceil} \in \mathbb{R}^{t \times d_k}, \\
    \bm{V}^{\lceil hg/n\rceil} &= \bm{XW}_v^{\lceil hg/n\rceil} \in \mathbb{R}^{t \times d_v}, \\
\end{aligned}
\end{equation}
where a common choice sets $d_k=d_v=d_{model}/h$, and if $g = h,$ this GQA setting reduces to standard multi-head attention. In the Llama-3 collection, a typical choice sets $g=8$ for KV sharing. The output of head $s$ is then computed as:
\begin{equation}
\begin{aligned}
    \bm{O}^{h} = \mathrm{Softmax}&\Bigl(
        \tfrac{\bm{Q}^{h}\,(\bm{K}^{\lceil hg/n\rceil})^\top}{\sqrt{d_k}}
      \Bigr)\,
      \bm{V}^{\lceil hg/n\rceil},
    \\
    \bm{O}^{h} &\in \mathbb{R}^{t \times d_v}.  
\end{aligned}
\end{equation}

\subsection{Task Head Parameters}
\label{sec:downstream_params}
Downstream task-related head parameters are critical components in LLMs that directly influence performance on specific tasks. Inspired by mechanistic interpretability studies \citep{voita2019analyzing,zhao2024explainability,lindner2023tracrcompiledtransformerslaboratory}, we define these task head parameters as those whose ablation results in a significant degradation of task-specific performance. Formally, given a downstream task $\mathcal{T}$ with evaluation metric $p$, we quantify the impact of ablating a head parameter $\theta^h$ by the relative performance drop:
\begin{equation}
\Delta p(\theta^{h}) = p\left(\theta^{\mathcal{M}}; \mathcal{T}\right) - p\left(\theta^{\mathcal{M}} \setminus \theta^{h}; \mathcal{T}\right),
\end{equation}
where $\theta^{\mathcal{M}}$ denotes the original model parameters, and $\theta^{\mathcal{M}} \setminus \theta^{h}$ represents the model after ablating head parameters $\theta^{h}$. The downstream task-related head parameters are then identified as the Top-K heads that maximize this performance drop:
\begin{equation}\label{eq:task_params}
\Theta_{\mathcal{T},K} = \text{Top-K} \left\{ \theta^{\mathcal{T}} : \operatorname*{\arg\!\max}_{\theta^{h} \in \theta^{\mathcal{M}}} \Delta p(\theta^{h}) \right\}.
\end{equation}

The ablation of $\Theta_{\mathcal{T},K}$ leads to a measurable decline in task performance, underscoring their importance.

\subsection{Attention Localization and Pruning Strategy (ALPS)}\label{sec:alps}
To localize task-sensitive attention heads that contribute to downstream task alignment, we introduce a heuristic search algorithm named \textit{\textbf{A}ttention \textbf{L}ocalization and \textbf{P}runing \textbf{S}trategy (\textbf{ALPS})}. The algorithm is divided into two stages: \textbf{Localizing} task-sensitive attention heads and \textbf{Pruning} non-critical heads in task alignment.

\paragraph{Localizing.} To identify task-sensitive attention heads, we introduce the \emph{parameter alignment distribution score} $s^{PAD}$, which is based on the Wasserstein-1 distance~\citep{vaserstein1969markov} between the weight matrices of a fine-tuned task model and its base counterpart, quantifying the variation in attention head parameters across different tasks.

Given a pre-trained model $\mathcal{M}_\mathcal{B}$ and its task fine-tuned version $\mathcal{M}_\mathcal{T}$, we analyze the static projection matrices of each attention head $h$. For head $h$ with parameters $\{\bm{W}_q^{h}, \bm{W}_k^{\lceil hg/n\rceil}, \bm{W}_v^{\lceil hg/n\rceil}\}$, the projection matrix is computed as:
\begin{equation}
    \bm{W}_o^{h} = \bm{W}_q^{h} \bm{W}_k^{\lceil hg/n\rceil\top} \bm{W}_v^{\lceil hg/n\rceil} \in \mathbb{R}^{d \times d},
\end{equation}
which captures the transformation behavior of one head\citep{kobayashi2020attention}. 

To measure task-induced parameter shifts, we first convert $\bm{W}_o^{(s)}$ into a probability distribution via tempered softmax:
\begin{equation}
    \bm{P}^{h} = \mathrm{Softmax}\left( \tfrac{\bm{W}_o^{h}}{\tau} \right), \quad \tau > 0,
\end{equation}
where $\tau$ controls distribution granularity, and we set $\tau$ to 1 for simplicity. The Wasserstein-1 distance between base and task distributions then quantifies head sensitivity:
\begin{equation}\label{eq:s^pad}
\begin{aligned}
    s_{h}^{PAD} &= W_1\left( \bm{P}_\mathcal{B}^{h}, \bm{P}_\mathcal{T}^{h} \right). \\
    &=\inf_{\gamma \in \Gamma(\bm{P}_\mathcal{B}^{h}, \bm{P}_\mathcal{T}^{h})} \mathbb{E}_{(x,y) \sim \gamma} \left[ \|x - y\| \right],
\end{aligned}
\end{equation}
where $\Gamma$ denotes all joint distributions with marginals $\bm{P}_\mathcal{B}^{h}$ and $\bm{P}_\mathcal{T}^{h}$, and we compute $s_{h}^{PAD}$ for each attention head. Results in Section~\ref{sec:ablation_study} show that heads with higher $s_{h}^{PAD}$ consistently achieve stronger task alignment and thus better performance, confirming that the magnitude of parameter variation reflects their importance for downstream tasks.

\begin{algorithm}[t]
\renewcommand{\algorithmicrequire}{\textbf{Input:}}
\renewcommand{\algorithmicensure}{\textbf{Output:}}
\caption{Attention Localization and Pruning Strategy (ALPS)}\label{alg:alps}
\begin{algorithmic}[1]
\REQUIRE $\mathcal{M}_\mathcal{B}$, $\mathcal{M}_\mathcal{T}$, $\mathcal{D}_\mathcal{T}$, retention ratio $r \in (0,1]$
\ENSURE $\mathcal{H}_r$, $\mathcal{M}_{ALPS}$
\STATE Let $\mathcal{H} \gets \{1,2,\dots, n\}$ \COMMENT{Set of all attention heads}
\FORALL{\(h \in \mathcal{H}\)}
    \STATE $\bm{P}_\mathcal{B}^{h} \gets\text{Softmax} (\bm{W}^{h}_{o,B})$
    \STATE $\bm{P}_\mathcal{T}^{h} \gets \text{Softmax} (\bm{W}^{h}_{o,T})$
    \STATE Compute $s^{PAD}_{h} \gets \mathcal{W}_1\Bigl(\bm{P}_\mathcal{B}^{h},\, \bm{P}_\mathcal{T}^{h}\Bigr)$
\ENDFOR
\STATE $K \gets \lceil r \cdot n \rceil$
\STATE $\mathcal{S}^{PAD} \gets \bigl\{ s_h^{PAD} \mid h \in \mathcal{H} \bigr\}$
\STATE $\mathcal{H}_\mathcal{T} \gets \operatorname{Top-K} \bigl( \mathcal{S}^{PAD} \bigr)$ \COMMENT{the set of heads with Top-$K$ $s_h^{PAD}$}
\FORALL{$\hat{h} \in \mathcal{H} \setminus \mathcal{H}_\mathcal{T}$}
    \STATE $\nabla_{\theta^{\hat{h}}} \mathcal{L} = 0 $
    \COMMENT{Ablate non-critical heads during fine-tuning}
\ENDFOR
\STATE Fine-tune model on $\mathcal{D}_\mathcal{T}$ to obtain $\mathcal{M}_{\text{ALPS}}$
\RETURN $\mathcal{H}_r$, $\mathcal{M}_{\text{ALPS}}$
\end{algorithmic}
\end{algorithm}

\begin{table*}[t]
    \centering
    \resizebox{\linewidth}{!}{%
        \begin{tabular}{cccccccccccl}
\toprule[1.2pt]
\multirow{2}{*}{\textbf{Model}} & \multirow{2}{*}{\textbf{Method}} & \multirow{2}{*}{\textbf{Attn\%}} & \multicolumn{2}{c}{\textbf{General}} & \multicolumn{2}{c}{\textbf{Math}} & \multicolumn{4}{c}{\textbf{Code}} & \multirow{2}{*}{\textbf{Avg.}} \\
\cmidrule(lr){4-5} \cmidrule(lr){6-7} \cmidrule(lr){8-11}
& & & IFEval & GPQA & GSM8K & MATH & HEval & HEval+ & MBPP & MBPP+ & \\

\midrule

\multirow{6}{*}{Llama-3.2-1B} & w/ full & 100\% & 34.53 & 18.75 & 17.82 & 4.60 & 45.12 & 39.08 & 28.63 & 23.78 & 26.54 \\
\cmidrule{2-12}
& w/o attn                              & 0\%   & 35.85 & 19.64 & 23.35 & 5.20 & 41.51 & 35.28 & \textbf{29.91} & \textbf{24.88} & 26.95 \footnotesize{\textcolor{teal}{$\uparrow$0.41}} \\
& Random                                & 10\%  & 33.29 & 25.89 & 22.82 & 5.02 & 41.51 & 37.22 & 28.12 & 24.13 & 27.25 \footnotesize{\textcolor{teal}{$\uparrow$0.71}}\\
& LC                                    & 10\%  & 32.15 & 27.88 & 21.73 & 4.81  & 42.05 & 38.07 & 27.03 & 23.88 & 27.20 \footnotesize{\textcolor{teal}{$\uparrow$0.66}} \\
& LoRA                    & \textasciitilde10\% & \textbf{36.75} & 26.14 & 23.67 & 5.92  & 42.34 & 37.99 & 28.92 & 24.10 & 28.22 \footnotesize{\textcolor{teal}{$\uparrow$1.68}} \\
\rowcolor{COLOR_MEAN}
\cellcolor[HTML]{FFFFFF} & \textit{\textbf{ALPS}}                  & 10\%  & 36.69 & \textbf{26.16} & \textbf{23.74} & \textbf{7.28} & \textbf{46.89} & \textbf{40.22} & 29.13 & 24.21 & \textbf{29.29 \footnotesize{\textcolor{teal}{$\uparrow$2.75}}} \\

\midrule

\multirow{6}{*}{Llama-3.2-3B} & w/ full & 100\% & 42.81 & 22.32 & 31.77 & 16.68 & 56.18 & 50.53 & 50.00 & 41.50 & 38.97 \\
\cmidrule{2-12}
& w/o attn                              & 0\%   & 40.29 & 24.33 & 33.21 & 13.76 & 55.52 & 50.62 & 48.71 & 39.72 & 38.27 \footnotesize{\textcolor{purple}{$\downarrow$0.70}} \\
& Random                                & 10\%  & 42.21 & 20.54 & 35.03 & 14.56 & 56.18 & 49.43 & 54.57 & 45.28 & 39.73 \footnotesize{\textcolor{teal}{$\uparrow$0.76}} \\
& LC                                    & 10\%  & 41.75 & 21.86 & \textbf{36.42} & 13.94 & 55.82 & 50.67 & 55.13 & 42.50 & 39.76 \footnotesize{\textcolor{teal}{$\uparrow$0.79}} \\
& LoRA                    & \textasciitilde10\% & 42.90 & 21.22 & 34.20 & 15.36 & 56.85 & 50.09 & \textbf{55.27} & \textbf{45.03} & 40.12 \footnotesize{\textcolor{teal}{$\uparrow$1.15}} \\
\rowcolor{COLOR_MEAN}
\cellcolor[HTML]{FFFFFF} & \textit{\textbf{ALPS}}                  & 10\%  & \textbf{44.96} & \textbf{24.33} & 34.27 & \textbf{17.12} & \textbf{58.28} & \textbf{51.67} & 52.21 & 42.78 & \textbf{40.70 \footnotesize{\textcolor{teal}{$\uparrow$1.73}}} \\

\midrule

\multirow{6}{*}{Llama-3.1-8B} & w/ full & 100\% & 51.20 & 25.22 & 63.00 & 20.52 & 69.50 & 64.28 & 63.21 & 52.38 & 51.16 \\
\cmidrule{2-12}
& w/o attn                              & 0\%   & 40.41 & 26.16 & 64.06 & 15.48 & 70.72 & \textbf{65.88} & 63.27 & 52.55 & 49.82 \footnotesize{\textcolor{purple}{$\downarrow$1.34}}\\
& Random                                & 10\%  & 49.52 & 26.34 & 62.40 & 23.52 & 70.12 & 63.39 & 59.00 & 49.21 & 50.44 \footnotesize{\textcolor{purple}{$\downarrow$0.72}} \\
& LC                                    & 10\%  & 48.67 & 27.10 & 63.83 & 22.47 & 70.68 & 63.05 & 58.28 & 49.79 & 50.48 \footnotesize{\textcolor{purple}{$\downarrow$0.68}} \\
& LoRA                    & \textasciitilde10\% & 50.02 & 26.82 & 62.92 & \textbf{24.07} & 72.50 & 63.89 & 59.55 & 49.69 & 51.18 \footnotesize{\textcolor{teal}{$\uparrow$0.02}} \\
\rowcolor{COLOR_MEAN}
\cellcolor[HTML]{FFFFFF} & \textit{\textbf{ALPS}}                  & 10\%  & \textbf{51.33} & \textbf{27.29} & \textbf{64.13} & 22.48 & \textbf{72.68} & 65.03 & \textbf{63.67} & \textbf{52.88} & \textbf{52.41 \footnotesize{\textcolor{teal}{$\uparrow$1.25}}} \\

\bottomrule[1.2pt]

\end{tabular}
    }
    \caption{Performance of our method compared to other baselines with Llama-3.2-1B, Llama-3.2-3B, and Llama-3.1-8B in general, math, and code tasks. The best results are highlighted in \textbf{bold}. {\footnotesize{\textcolor{purple}{$\downarrow$}}}{\footnotesize{\textcolor{teal}{$\uparrow$}}} indicates change relative to the w/ full baseline. Note that each task uses a separately fine-tuned model trained on its respective dataset. \textbf{ALPS} achieves better performance while updating only \textbf{10\%} of attention parameters, highlighting its efficiency.
    }
    \label{tab:main_table}
\end{table*}

\paragraph{Pruning.} After obtaining a set of $\mathcal{S}^{PAD}= \bigl\{ s_h^{PAD} \mid h \in \mathcal{H} \bigr\}$, where $\mathcal{H} = \{1, \dots, n\}$ denote all attention heads in the model, we select a group of task-sensitive heads $\mathcal{H}_{\mathcal{T}}$ with $\operatorname{Top-K} \bigl( \mathcal{S}^{PAD} \bigr)$ to retain. During task fine-tuning, the remaining heads will be pruned by freezing gradients of these heads, which masks their parameter updates to 0. Specifically, for all attention heads, only parameters associated with $\mathcal{H}_\mathcal{T}$ are updated:
\begin{equation}\label{eq:gradient_update}
    \nabla_{\theta^{h}} \mathcal{L} = 
    \begin{cases} 
    \nabla_{\theta^{h}} \mathcal{L} & \text{if } h \in \mathcal{H}_S \\
    0 & \text{otherwise}
    \end{cases},
\end{equation}
where $\theta^{h} = \{\bm{W}_q^{(h)}, \bm{W}_k^{(h)}, \bm{W}_v^{(h)}\}$. This gradient masking ensures that non-sensitive heads $\mathcal{H} \setminus \mathcal{H}_\mathcal{T}$ retain pre-trained knowledge without interference, and the loss objective turns to:
\begin{equation}
\mathcal{L} = \mathbb{E}_{(x,y) \sim \mathcal{D}_{\mathcal{T}}} \left[ -\log p\left(y \mid x; \{\theta^h\}_{h \in \mathcal{H}_\mathcal{T}} \right) \right],
\end{equation}
where $\{\theta_s\}$ denotes the parameters of heads in $\mathcal{H}_S$, and $\mathcal{D}_\mathcal{T}$ represents the downstream task data. This approach reduces optimization redundancy by focusing updates on heads critical to $\mathcal{T}$.

The overall procedure is detailed in Algorithm~\ref{alg:alps}, and the final output of the algorithm is the set of sensitive attention heads $\mathcal{H}_\mathcal{T}$ with modified fine-tuned model $\mathcal{M}_{ALPS}$.

\section{Experiments}\label{sec:experiments}
\subsection{Setup}\label{sec:setup}

\begin{table*}[h!]
    \centering
    \resizebox{\linewidth}{!}{%
        \begin{tabular}{cccccccccccc}
\toprule[1.2pt]

\multirow{2}{*}{\textbf{Model}} & \multirow{2}{*}{\textbf{Method}} & \multirow{2}{*}{\textbf{Attn\%}} & \multicolumn{2}{c}{\textbf{General}} & \multicolumn{2}{c}{\textbf{Math}} & \multicolumn{4}{c}{\textbf{Code}} & \multirow{2}{*}{\textbf{Avg.}} \\
\cmidrule(lr){4-5} \cmidrule(lr){6-7} \cmidrule(lr){8-11}
& & & IFEval & GPQA & GSM8K & MATH & HEval & HEval+ & MBPP & MBPP+ & \\

\midrule

\multirow{4}{*}{Llama-3.2-1B} 
& C.S. & 10\% 
  & 28.12 & 17.97 & 18.18 & 4.15  & 28.22 & 23.97 & 18.89 & 13.67 & 19.15 \\
& Eu.  & 10\% 
  & 34.05 & 24.68 & 21.91 & 6.89  & 44.55 & 39.12 & 27.14 & 22.54 & 27.61 \\
& KL   & 10\% 
  & 35.22 & \textbf{26.45} & \textbf{23.95} & 7.05  & 45.83 & \textbf{40.45} & 28.72 & \textbf{24.55} & 29.03 \\
\rowcolor{COLOR_MEAN}
\cellcolor[HTML]{FFFFFF} & \textbf{$s^{PAD}$}& 10\% & \textbf{36.69} & 26.16 & 23.74 & \textbf{7.28} & \textbf{46.89} & 40.22 & \textbf{29.13} & 24.21 & \textbf{29.29} \\

\midrule

\multirow{4}{*}{Llama-3.2-3B} 
& C.S. & 10\% 
  & 32.57 & 19.88 & 22.05 & 10.67  & 46.12 & 38.83 & 38.97 & 30.25 & 29.92 \\
& Eu.  & 10\% 
  & 42.36 & 22.45 & 32.11 & 16.23  & 55.67 & 49.25 & 49.82 & 40.15 & 38.51 \\
& KL   & 10\% 
  & 43.78 & \textbf{24.50} & 33.14 & 16.78  & 56.89 & 50.33 & 50.47 & 41.12 & 39.63 \\
\rowcolor{COLOR_MEAN}
\cellcolor[HTML]{FFFFFF} & \textbf{$s^{PAD}$} & 10\%  & \textbf{44.96} & 24.33 & \textbf{34.27} & \textbf{17.12} & \textbf{58.28} & \textbf{51.67} & \textbf{52.21} & \textbf{42.78} & \textbf{40.70}  \\

\midrule

\multirow{4}{*}{Llama-3.1-8B} 
& C.S. & 10\% 
  & 41.85 & 22.47 & 48.22 & 19.17  & 61.45 & 52.93 & 55.34 & 49.05 & 43.81 \\
& Eu.  & 10\% 
  & 48.94 & 26.03 & 62.78 & 20.84  & 69.12 & 63.47 & 61.02 & 50.37 & 50.32 \\
& KL   & 10\% 
  & 50.12 & 26.75 & 63.91 & 21.26  & \textbf{73.05} & 64.12 & 62.15 & 51.43 & 51.60 \\
\rowcolor{COLOR_MEAN}
\cellcolor[HTML]{FFFFFF} & \textbf{$s^{PAD}$} & 10\% & \textbf{51.33} & \textbf{27.13} & \textbf{64.13} & \textbf{22.48} & 72.68 & \textbf{65.03} & \textbf{63.67} & \textbf{52.88} & \textbf{52.41} \\

\bottomrule[1.2pt]

\end{tabular}
    }
    \caption{Ablation study on metrics for \NAME with 10\% of attention heads. \textbf{Bold} indicates the best performance.}
    \label{tab:metric_ablation}
\end{table*}
\begin{table*}[h!]
    \centering
    \resizebox{0.9\linewidth}{!}{%
        \begin{tabular}{ccccccccccl}
\toprule[1.2pt]

\multirow{2}{*}{\textbf{Model}} & \multirow{2}{*}{\textbf{Ratio}} & \multicolumn{2}{c}{\textbf{General}} & \multicolumn{2}{c}{\textbf{Math}} & \multicolumn{4}{c}{\textbf{Code}} & \multirow{2}{*}{\textbf{Avg.}} \\
\cmidrule(lr){3-4} \cmidrule(lr){5-6} \cmidrule(lr){7-10}
& & IFEval & GPQA & GSM8K & MATH & HEval & HEval+ & MBPP & MBPP+ & \\

\midrule

\multirow{6}{*}{Llama-3.2-1B} 
& 100\% 
    & 34.53 & 18.75 & 17.82 & 4.60 & 45.12 & 39.08 & 28.63 & 23.78 & 26.54 \\
\cmidrule{2-11}
& 70\%
    & 31.80 & 16.91 & 20.63 & 6.35 & 44.20  & 39.50 & 28.40 & 24.90 & 26.59 \footnotesize{\textcolor{teal}{$\uparrow$0.05}} \\
& 50\% 
    & 33.46 & 18.07 & 21.08 & 7.15  & 43.95 & 39.69 & 28.92 & 24.64 & 27.12 \footnotesize{\textcolor{teal}{$\uparrow$0.58}} \\
& 30\%
    & \textbf{37.12} & 25.98 & 22.56 & 7.12  & 46.34 & 39.84 & 28.95 & 24.05 & 29.00 \footnotesize{\textcolor{teal}{$\uparrow$2.46}} \\
\rowcolor{COLOR_MEAN}
\cellcolor[HTML]{FFFFFF}
& \textbf{10\%} 
    & 36.69 & \textbf{26.16} & \textbf{23.74} & \textbf{7.28} & \textbf{46.89} & \textbf{40.22} & 29.13 & 24.21 & \textbf{29.29 \footnotesize{\textcolor{teal}{$\uparrow$2.75}}} \\
& 0\% 
    & 35.85 & 19.64 & 23.35 & 5.20 & 41.51 & 35.28 & \textbf{29.91} & \textbf{24.88} & 26.95 \footnotesize{\textcolor{teal}{$\uparrow$0.41}} \\

\midrule

\multirow{6}{*}{Llama-3.2-3B} 
& 100\% & 42.81 & 22.32 & 31.77 & 16.68 & 56.18 & 50.53 & 50.00 & 41.50 & 38.97 \\
\cmidrule{2-11}
& 70\% 
    & 39.83 & 21.91 & 31.63 & 15.35 & 56.21  & 51.55 & 49.40 & 39.97 & 38.23 \footnotesize{\textcolor{purple}{$\downarrow$0.74}} \\
& 50\% 
    & 40.93 & 22.18 & 32.17 & 16.10 & 55.91 & 49.72 & 50.05 & 40.33 & 38.42 \footnotesize{\textcolor{purple}{$\downarrow$0.55}} \\
& 30\% 
    & 43.53 & 23.46 & 33.91 & 16.54 & 57.28 & 51.69 & 51.61 & 41.93 & 39.99 \footnotesize{\textcolor{teal}{$\uparrow$1.02}} \\
\rowcolor{COLOR_MEAN}
\cellcolor[HTML]{FFFFFF}
& \textbf{10\%} 
    & \textbf{44.96} & \textbf{24.33} & \textbf{34.27} & \textbf{17.12} & \textbf{58.28} & \textbf{51.67} & \textbf{52.21} & \textbf{42.78} & \textbf{40.70 \footnotesize{\textcolor{teal}{$\uparrow$1.73}}} \\
& 0\% 
    & 40.29 & 24.33 & 33.21 & 13.76 & 55.52 & 50.62 & 48.71 & 39.72 & 38.27 \footnotesize{\textcolor{purple}{$\downarrow$0.70}} \\

\midrule

\multirow{6}{*}{Llama-3.1-8B} 
& 100\% 
    & 51.20 & 25.22 & 63.00 & 20.52 & 69.50 & 64.28 & 63.21 & 52.38 & 51.16 \\
\cmidrule{2-11}
& 70\%
    & 50.80 & 24.91 & 62.63 & 20.35 & 69.20 & 61.50 & 63.40 & 51.90 & 50.59 \footnotesize{\textcolor{purple}{$\downarrow$0.57}} \\
& 50\% 
    & \textbf{52.55} & \textbf{27.28} & 61.11 & 20.95 & 69.69 & 63.35 & 61.71 & 51.02 & 50.96 \footnotesize{\textcolor{purple}{$\downarrow$0.20}} \\
& 30\% 
    & 51.85 & 26.47 & 62.72 & 22.15 & 71.15 & 64.55 & 62.88 & 52.23 & 51.62 \footnotesize{\textcolor{teal}{$\uparrow$0.46}} \\
\rowcolor{COLOR_MEAN}
\cellcolor[HTML]{FFFFFF}
& \textbf{10\%} 
    & 51.33 & 27.13 & \textbf{64.13} & \textbf{22.48} & \textbf{72.68} & \textbf{65.03} & \textbf{63.67} & \textbf{52.88} & \textbf{52.41 \footnotesize{\textcolor{teal}{$\uparrow$1.25}}} \\
& 0\% 
    & 40.41 & 26.16 & 64.06 & 15.48 & 70.72 & 64.88 & 63.27 & 52.55 & 49.69 \footnotesize{\textcolor{purple}{$\downarrow$1.47}} \\

\bottomrule[1.2pt]

\end{tabular}
    }
    \caption{Ablation study on attention head selection ratio for \NAME. The best results are highlighted in \textbf{bold}. {\footnotesize{\textcolor{purple}{$\downarrow$}}}{\footnotesize{\textcolor{teal}{$\uparrow$}}} indicates change relative to the w/ full baseline.
    }
    \label{tab:ratio_ablation}
\end{table*}

\paragraph{Datasets.} We evaluate our approach on three representative datasets covering general, math, and code tasks: UltraChat~\citep{ding2023enhancing}, MathInstruct~\citep{yue2023mammoth}, and MagiCoder~\citep{wei2023magicoder}. To assess the transferability of our method, we further conduct experiments on Alpaca~\citep{alpaca}, Camel-math~\citep{li2023camel}, and CodeAlpaca~\citep{codealpaca}. Details are listed in Appendix~\ref{appx_sec:datasetsnbenchmarks}.

\paragraph{Models.} We conduct our method on the Llama-3 series models (Llama-3.2-1B, Llama-3.2-3B, and Llama-3.18B), all pre-trained with GQA for improved efficiency. Each model shares key and value projections across 8 heads ($g$=8), while queries remain head-specific. Thus, when computing $s^{PAD}$, each key-value pair corresponds to 8 query heads.

\paragraph{Evaluation.} We use lm-eval~\citep{eval-harness,llama-reciepe} to evaluate general capabilities in instruction following, reasoning, and dialogue, as well as mathematical abilities. For code generation, we employ EvalPlus~\citep{evalplus}. All evaluations follow the official implementations, with details on benchmarks and metrics provided in Appendix~\ref{appx_sec:datasetsnbenchmarks}.

\paragraph{Baseline.} To evaluate the effectiveness of \NAME, we compare it against the following baselines. \textbf{w/ full}: standard full supervised fine-tuning where all attention parameters are updated. \textbf{w/o attn}: freeze all attention heads by preventing updates to the QKV weight matrices. \textbf{Random}: selects a fixed proportion of random attention heads for training. \textbf{LC}: layer consistency selection, extending random by ensuring a uniform distribution of selected heads across layers, maintaining fairness in head allocation. \textbf{LoRA}~\citep{hu2021lora}: a widely used parameter-efficient fine-tuning method that freezes all original weights and updates only a low-rank decomposition of weight matrices. We set the LoRA rank to 128 and only to QKV matrices. For consistency in head indexing, all head selections follow the query index. For a Q index of 20, the K and V indices are determined by $\lceil 20 * g/h \rceil$, where $g$ is the number of KV groups, and $h$ is the total number of heads.

\paragraph{Implementation Details.}
We optimize our models using the AdamW \citep{loshchilov2017decoupled} optimizer with hyperparameters $\beta_1=0.9$ and $\beta_2=0.999$, and a weight decay of 0.1. A cosine decay schedule is employed, gradually reducing the learning rate to 10\% of its initial value throughout training, following a linear warmup ratio of 0.1. We set the effective batch size to 128 and initialize the learning rate at $2 \times 10^{-5}$. All models are trained for 3 epochs using mixed precision (fp16) on 8 A100 GPUs.

\subsection{Main Results}

Table~\ref{tab:main_table} presents the main results comparing \NAME against other baselines across three model scales (1B, 3B, and 8B) and three downstream tasks (general, math, and code). Note that each task uses a separately fine-tuned model trained on its respective dataset, that is, the results of general tasks are from the general fine-tuned model, while the results of math and code are from math and code fine-tuned models, respectively. Notably, our method consistently outperforms all baselines, showing a 2.75\% improvement over the \emph{w/ full} baseline on the 1B model and maintains a 1.25\% gain at the 8B scale. This indicates the redundancy among attention heads in downstream alignment. In contrast, the \emph{w/o attn} settings fall significantly behind the \emph{w/ full} baseline, highlighting the critical role of attention heads in downstream alignment. Furthermore, results from \emph{Random}, \emph{LC}, \emph{LoRA}, and our method indicate that selecting a subset of heads can indeed enhance alignment efficiency. However, how these heads are chosen remains pivotal, and selecting more task-sensitive attention heads can outperform training on all heads, underscoring the importance of precise head identification for maximizing both efficiency and performance.

\begin{figure}[ht]
    \centering
    \includegraphics[width=0.9\linewidth]{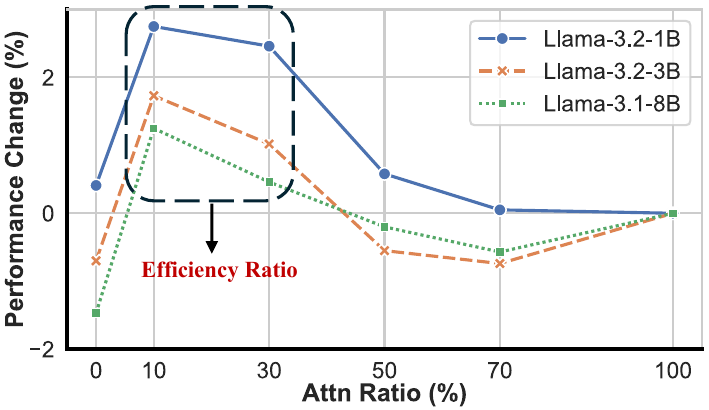}
    \caption{Ablation study on attention head selection ratio. The efficiency ratio indicates that activating 10\% and 30\% of attention heads yields better performance than full fine-tuning across three models.}
    \label{fig:ratio_ablation}
\end{figure}

\subsection{Ablations Study}\label{sec:ablation_study}
\paragraph{Head Metric Ablation.}
Table~\ref{tab:metric_ablation} compares four metrics: \emph{C.S.} (cosine similarity), \emph{Eu.} (Euclidean distance), \emph{KL} (Kullback-Leibler Divergence~\citep{kullback1951information}), and our proposed \emph{parameter alignment distribution score} $s^{PAD}$, for identifying task-sensitive attention heads. Notably, \emph{C.S.} yields the weakest performance, likely because it focuses on overall matrix similarity, which can overlook significant local shifts tied to specific task requirements. In contrast, across all model scales (1B, 3B, 8B) and tasks (general, math, code), our proposed metric $s^{PAD}$ consistently achieves the highest average scores, indicating its effectiveness in capturing the most critical distribution shifts between the base and task fine-tuned models.

\paragraph{Head Ratio Ablation.}
The ablation study on the proportion of attention heads selected for our method ranging from \emph{0\%} to \emph{100\%}, is presented in Table~\ref{tab:ratio_ablation} and Figure~\ref{fig:ratio_ablation}. We observe that both \emph{10\%} and \emph{30\%} ratios consistently yield strong performance across all model scales and tasks, as highlighted by the efficiency ratio region in Figure~\ref{fig:ratio_ablation}. Notably, updating 10\% of the heads already achieves comparable to or better results than higher ratios, and strikes a better trade-off between performance and computational cost by reducing the number of trainable parameters. Consequently, we adopt \emph{10\%} as our default setting for improved efficiency.

\begin{figure}[H]
    \centering
    \includegraphics[width=0.9\linewidth]{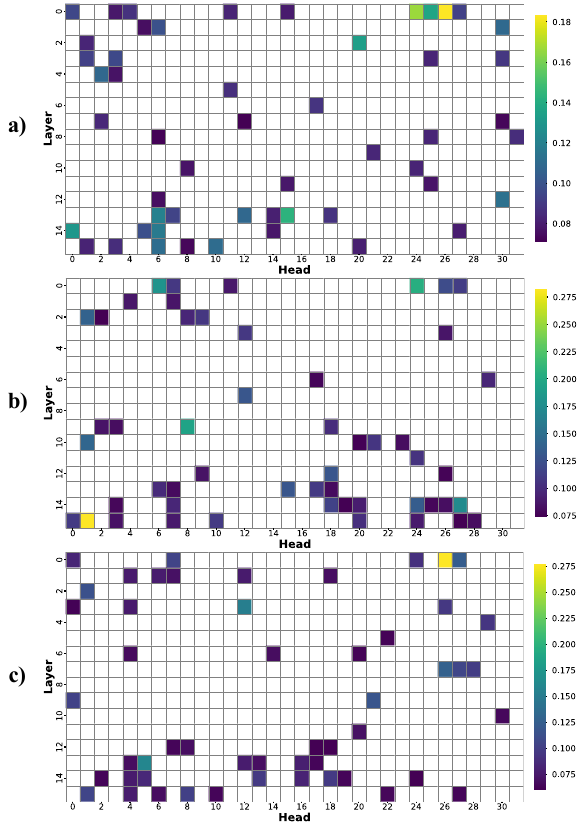}
    \caption{Heatmaps of task-sensitive attention heads (10\%) selected by our method on Llama-3.2-1B for \textbf{a)} general, \textbf{b)} math, and \textbf{c)} code tasks. The color intensity indicates the value of $s^{PAD}$ of each head.}
    \label{fig:heatmap}
\end{figure}

\subsection{Head Impact Analysis}
Figure~\ref{fig:heatmap} shows heatmaps of the top 10\% attention heads identified by our method across three tasks, full heatmaps are listed in Appendix~\ref{appx_sec:heatmaps}. Each grid cell corresponds to a layer-head combination, and the color intensity reflects the importance of the head. We observe that general and code tasks exhibit notable overlap in their most sensitive heads, potentially due to the presence of extensive natural language instructions in the code datasets. Meanwhile, math reveals a more distinct pattern, suggesting specialized attention requirements for mathematical reasoning. Some heads appear consistently important across all three tasks, indicating partial head sharing that aligns with similarities in data distributions as illustrated in Appendix~\ref{appx_sec:datasetsnbenchmarks}. These findings highlight that our method effectively identifies both task-specific and task-agnostic heads, shedding light on how different downstream objectives influence attention-head specialization.

\begin{table}[t]
    \centering
    \resizebox{\linewidth}{!}{%
        \begin{tabular}{cccccl}
\toprule[1.2pt]

\textbf{Model} & \textbf{Method} & \textbf{General} & \textbf{Math} & \textbf{Code} & \textbf{Avg.} \\

\midrule

\multirow{4}{*}{Llama-3.2-1B} 
& w/ full 
    & 14.28 & 5.87 & 22.28 & 14.14 \\
\cmidrule(lr){2-6}
& w/o attn 
    & 19.73 & 8.39 & 19.98 & 16.03 \footnotesize{\textcolor{teal}{$\uparrow$1.89}} \\
& Random 
    & 21.80 & 7.91 & 19.63 & 16.45 \footnotesize{\textcolor{teal}{$\uparrow$2.31}} \\
\rowcolor{COLOR_MEAN}
\cellcolor[HTML]{FFFFFF} 
& \textit{\textbf{ALPS}} 
    & \textbf{23.43} & \textbf{11.51} & \textbf{24.11} & \textbf{19.68 \footnotesize{\textcolor{teal}{$\uparrow$5.54}}} \\

\midrule

\multirow{4}{*}{Llama-3.2-3B} 
& w/ full 
    & 21.57 & 15.23 & 38.55 & 25.12 \\
\cmidrule(lr){2-6}
& w/o attn 
    & 22.31 & 13.49 & 35.72 & 23.84 \footnotesize{\textcolor{purple}{$\downarrow$1.28}} \\
& Random 
    & 22.38 & 17.80 & \textbf{41.37} & 27.18 \footnotesize{\textcolor{teal}{$\uparrow$2.06}} \\
\rowcolor{COLOR_MEAN}
\cellcolor[HTML]{FFFFFF} 
& \textit{\textbf{ALPS}} 
    & \textbf{24.65} & \textbf{18.70} & 40.24 & \textbf{27.86 \footnotesize{\textcolor{teal}{$\uparrow$2.74}}} \\

\midrule

\multirow{4}{*}{Llama-3.1-8B} 
& w/ full 
    & 27.21 & 28.76 & 42.34 & 32.77 \\
\cmidrule(lr){2-6}
& w/o attn 
    & 18.29 & 25.77 & \textbf{46.11} & 30.06 \footnotesize{\textcolor{purple}{$\downarrow$3.38}} \\
& Random 
    & 27.93 & \textbf{32.96} & 40.43 & 33.77 \footnotesize{\textcolor{teal}{$\uparrow$1.00}} \\
\rowcolor{COLOR_MEAN}
\cellcolor[HTML]{FFFFFF} 
& \textit{\textbf{ALPS}} 
    & \textbf{28.21} & 31.31 & 44.57 & \textbf{34.67 \footnotesize{\textcolor{teal}{$\uparrow$1.90}}} \\

\bottomrule[1.2pt]

\end{tabular}
    }
    \caption{Performance of identified heads with a new set of general, math, and code datasets. The best results are highlighted in \textbf{bold}. {\footnotesize{\textcolor{purple}{$\downarrow$}}}{\footnotesize{\textcolor{teal}{$\uparrow$}}} indicates change relative to the baseline. 
    Our method consistently outperforms other methods, demonstrating the transferability of selected task heads. Full details are listed in Appendix~\ref{sec:full_tables}.
    }
    \label{tab:head_transfer}
\end{table}

\begin{figure}[H]
    \centering
    \includegraphics[width=0.9\linewidth]{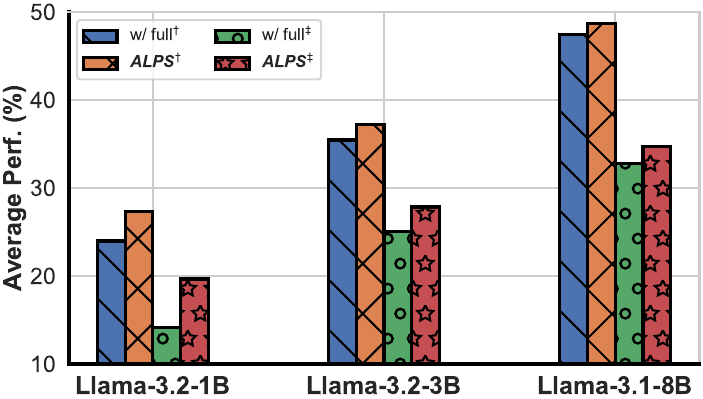}
    \caption{Evaluation on transferability of \textbf{ALPS} with a new set of general, math, and code datasets. Results with $\dagger$ correspond to results from Table~\ref{tab:main_table}, while $\ddagger$ represents performance from Table~\ref{tab:head_transfer}.}
    \label{fig:head_transfer}
\end{figure}

\subsection{Does Heads Transfer?}

To assess whether ALPS-selected heads generalize across different datasets within the same task domain, we evaluate them on a new set of general, math, and code datasets, as illustrated in Section~\ref{sec:setup}. Table~\ref{tab:head_transfer} and Figure~\ref{fig:head_transfer} show that our method consistently outperforms all baselines, indicating the transferability of the task-sensitive attention heads identified by \NAME that can be reused effectively on new datasets within the same tasks. This highlights the potential of our method for reducing alignment costs without sacrificing performance.

\subsection{Fewer Heads Avoid Forgetting}

Table~\ref{tab:forgetting} presents results on MMLU and ARC-C, two commonly used benchmarks to assess general knowledge, making them vulnerable to knowledge forgetting when the alignment dataset is limited in quality or quantity~\cite{chang2024survey}. The \emph{w/ full} baseline shows a noticeable drop in performance, reflecting the loss of pre-trained knowledge. In contrast, \NAME, which restricts updates to a small fraction of attention heads, better preserves this general knowledge and consistently outperforms both \emph{w/ full}, \emph{w/o attn}, and \emph{Random} baselines. By limiting the number of trained heads that introduce sparsity, our method reduces overfitting to the smaller task-specific dataset, thereby mitigating knowledge forgetting from the pre-training phase.

\begin{table}[t]
    \centering
    \resizebox{0.8\linewidth}{!}{%
        \begin{tabular}{ccccc}
\toprule[1.2pt]

\textbf{Model} & \textbf{Method} & \textbf{MMLU} & \textbf{ARC-C} & \textbf{Avg.} \\

\midrule

\multirow{5}{*}{Llama-3.2-1B} 
& vanilla 
    & 32.2 & 32.8 & 32.5 \\
\cmidrule(lr){2-5}
& w/ full 
    & 28.14 & 26.95 & 27.55  \\
& w/o attn 
    & 27.80 & 23.18 & 25.49 \\
& Random 
    & 27.86 & 26.27 & 27.07 \\
\rowcolor{COLOR_MEAN}
\cellcolor[HTML]{FFFFFF} 
& \textit{\textbf{ALPS}} 
    & \textbf{29.88} & \textbf{28.73} & \textbf{29.31} \\

\midrule

\multirow{5}{*}{Llama-3.2-3B} 
& vanilla 
    & 58 & 69.1 & 63.55 \\
\cmidrule(lr){2-5}
& w/ full 
    & 44.22 & 50.82 & 47.52 \\
& w/o attn 
    & 46.16 & 46.95 & 46.56 \\
& Random 
    & 45.55 & 47.30 & 46.43 \\
\rowcolor{COLOR_MEAN}
\cellcolor[HTML]{FFFFFF} 
& \textit{\textbf{ALPS}} 
    & \textbf{48.83} & \textbf{52.37} & \textbf{50.60} \\

\midrule

\multirow{5}{*}{Llama-3.1-8B} 
& vanilla 
    & 66.7 & 79.7 & 73.2 \\
\cmidrule(lr){2-5}
& w/ full 
    & 55.40 & 60.34 & 57.87  \\
& w/o attn 
    & 22.95 & 49.01 & 35.98  \\
& Random 
    & 54.59 & 62.92 & 58.76  \\
\rowcolor{COLOR_MEAN}
\cellcolor[HTML]{FFFFFF} & 
\textit{\textbf{ALPS}} 
    & \textbf{57.87} & \textbf{64.29} & \textbf{61.08}  \\

\bottomrule[1.2pt]

\end{tabular}
    }
    \caption{Performance comparison on two common general benchmarks between our method and baselines with the vanilla models. Full details are listed in Appendix~\ref{sec:full_tables}.}
    \label{tab:forgetting}
\end{table}

\section{Conclusion}
This work presents \NAME, an approach that leverages model weight parameters to localize and prune task-sensitive attention heads, leading to more efficient LLM alignment while avoiding data dependency. Extensive experiments on general, math, and code tasks demonstrate that our method consistently outperforms baselines, exhibits head transferability, and mitigates knowledge forgetting. Overall, \NAME offers a promising perspective on efficient LLM alignment and paves the way for further research into parameter-efficient alignment strategies that capitalize on the role of attention heads. 

\section{Limitation}
While \NAME demonstrates promising improvements in efficiency and task performance, our investigation into its transferability and its ability to mitigate knowledge forgetting remains preliminary. Due to training overhead constraints, our evaluations have been limited to a select set of downstream tasks and datasets. Moreover, further investigation is needed to elucidate the underlying mechanism by which \NAME preserves pre-trained knowledge. Despite these limitations, our approach still effectively avoids data dependency and highlights a promising direction for future research in efficient LLM alignment.

\section{Acknowledgment}
This work is mainly supported by the National Key Research and Development Program of China (No. 2022YFB3304100), and partially by the Fundamental Research Funds for the Central Universities (226-2024-00049).

\bibliography{custom}

\appendix

\section{Datasets and Benchmarks}
\label{appx_sec:datasetsnbenchmarks}

\paragraph{Datasets.} Table~\ref{tab:dataset_info} provides an overview of the datasets used in our experiments. Each dataset consists of $\langle$\textit{instruction, answer}$\rangle$ pairs, which serve as the basis for supervised fine-tuning. These datasets span diverse domains, ensuring a comprehensive evaluation of our approach across general, math, and code tasks

\begin{table}[h!]
    \centering
    \resizebox{\linewidth}{!}{%
        \begin{tabular}{ccr}
\toprule[1.2pt]
\textbf{Name} & \textbf{Task} & \textbf{\# Samples} \\ \midrule
UltraChat~\citep{ding2023enhancing} & general & 200k \\
MathInstruct~\citep{yue2023mammoth} & math & 262k \\
Magicoder~\citep{wei2023magicoder} & code & 110k \\
Alpaca~\citep{alpaca} & general & 52k \\
Camel-math~\citep{li2023camel} & math & 50k \\
CodeAlpaca~\citep{codealpaca} & code & 20k \\
\bottomrule[1.2pt]
\end{tabular}
    }
    \caption{Overview of the datasets employed in our experiments, including dataset name, task domain, and number of samples.}
    \label{tab:dataset_info}
\end{table}

\begin{figure}[H]
    \centering
    \includegraphics[width=0.8\linewidth]{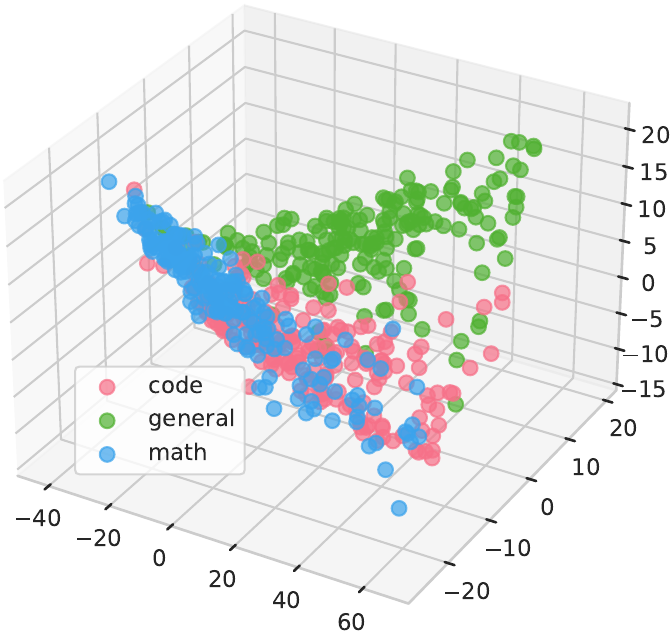}
    \caption{3D visualization of data distribution for UltraChat, MathInstruct, and Magicoder. All data is embedded using the Llama-3.1-8B model.}
    \label{fig:data_distribution}
\end{figure}

\paragraph{Data Distribution.} Figure~\ref{fig:data_distribution} shows a 3D visualization of the data distribution for UltraChat, MathInstruct, and Magicoder, as embedded by the Llama-3.1-8B model. Notably,  the code and math datasets exhibit some overlaps, whereas the general dataset displays a distinct distribution, reflecting domain-specific characteristics.

\paragraph{Evaluations.} We evaluate model performance using lm-eval~\citep{eval-harness,llama-reciepe} to assess general capabilities such as instruction following, reasoning, dialogue, and mathematical abilities, and EvalPlus~\citep{evalplus} to evaluate the capabilities of code generation. All settings are with the official report, and the details regarding the evaluation benchmarks, including the number of shots and metrics, are summarized in Table~\ref{tab:eval_info}, with further metric-specific details provided in Table~\ref{tab:eval_metric_info}

\begin{table*}[h!]
    \centering
    \begin{tabular}{cccc}
\toprule[1.2pt]
\textbf{Benchmark} & \textbf{Capability} & \textbf{\# Shots} & \textbf{Metric} \\ \midrule
\multirow{2}{*}{IFEval~\citep{zhou2023instruction}} & Instruction & \multirow{2}{*}{-} & Avg(Prompt/Instruction \\
& Following & & acc Loose/Strict) \\
GPQA~\citep{rein2023gpqa} & Reasoning & 0 & acc \\
GSM8K~\citep{cobbe2021training} & Math & 8 & em\_maj1@1 \\
MATH~\citep{hendrycks2021measuring} & Math & 0 & final\_em \\
Humaneval~\cite{chen2021evaluating} & Code & 0 & pass@1 \\
Humaneval+~\citep{evalplus} & Code & 0 & pass@1 \\
MBPP~\citep{austin2021program} & Code & 0 & pass@1 \\
MBPP+~\citep{evalplus} & Code & 0 & pass@1 \\
MMLU~\citep{hendrycks2020measuring} & General & 5 & macro\_avg/acc \\
ARC-C~\citep{clark2018think} & Reasoning & 0 & acc \\

\bottomrule[1.2pt]
\end{tabular}%

    \caption{Overview of benchmarks utilized in our experiments, detailing the assessed capability, number of shots, and evaluation metrics.}
    \label{tab:eval_info}
\end{table*}

\begin{table*}[h!]
    \centering
    \begin{tabular}{cc}
\toprule[1.2pt]
\textbf{Metric} & \textbf{Detail} \\ \midrule
macro\_avg/acc & \multirow{2}{*}{The mean accuracy across all classes or tasks} \\
(Macro Average Accuracy) & \\
\midrule
acc\_char & \multirow{2}{*}{Evaluates character-level correctness} \\
(Character-Level Accuracy) & \\
\midrule
em & Measures how often the model output, \\
(Exact Match) & exactly matches the reference answer \\
\midrule
f1 & Balances precision and recall, \\
(F1 Score) & particularly useful for imbalanced data \\
\midrule
\multirow{2}{*}{pass@1} & Assesses the correctness of generated code, \\
& on the first attempt \\
\midrule
\multirow{2}{*}{em\_maj1@1} & Measures exact-match on the first major attempt, \\
& especially in complex reasoning or math problems \\
\midrule
\multirow{2}{*}{final\_em} & The final exact-match score, \\
& commonly used in challenging benchmarks \\
\bottomrule[1.2pt]
\end{tabular}%


    \caption{Overview of evaluation metric details.}
    \label{tab:eval_metric_info}
\end{table*}

\section{Training Details}
\label{appx_sec:training}
For all experiments, we use the Alpaca~\citep{alpaca} template for fine-tuning across all datasets and models. Table~\ref{tab:train_template} shows the training template employed, which standardizes the input format and instruction style for all tasks. This consistency facilitates effective fine-tuning across diverse domains.

\begin{table*}[h!]
    \centering 
    \begin{tabular}{c|c}
\toprule[1.2pt]
\textbf{ Field} &  \textbf{Content}\\ \midrule
\multirow{2}{*}{System prompt} & Below is an instruction that describes a task. \\
& Write a response that appropriately completes the request. \\
\midrule
\multirow{2}{*}{User prompt} & \#\#\# Instruction: \{\{content\}\} \\
& \#\#\# Response: \\

\bottomrule[1.2pt]
\end{tabular}%

    \caption{Training template used for fine-tuning.}
    \label{tab:train_template}
\end{table*}

\section{Does \textbf{ALPS} Require a Full Fine-tuning Cycle to Obtain a Task-related Model?}
For better clarification, the task-related model in our method serves merely as a recipe to extract the parameter shifts needed for head identification, and \textbf{the primary focus of \textbf{ALPS} is on the identification of task-sensitive attention heads for subsequent parameter-efficient fine-tuning (PEFT)}, rather than on the process of obtaining a task-related model. For domains such as code and math, well-established task-specific models, e.g., CodeLlama~\cite{roziere2023codellama} and Qwen-2.5-Coder~\cite{hui2024qwen25coder}, are readily available and can be used directly in conjunction with the same size and architecture model, e.g., Llama-2~\cite{touvron2023llama2}, Qwen-2.5~\cite{qwen25technicalreport} to perform \textbf{ALPS} head identification, and thereby obviating the need for additional full fine-tuning. Since there is no widely recognized task fine-tuned model with the Llama-3.2 series, we then opted to perform full SFT to obtain a task-related baseline. Importantly, this choice was made to ensure a fair comparison and does not affect the core contribution of \textbf{ALPS}, which lies in its efficient head identification strategy. Furthermore, transferable experiments in Table~\ref{tab:head_transfer} demonstrate that once the task-sensitive heads are identified, they remain transferable across subsequent fine-tuning cycles. This one localization enables significant reductions in training cost for all downstream tasks, which serves as a key strength of \textbf{ALPS}.

To further support our claim that ALPS does not necessarily involve a full fine-tuning cycle to obtain a task-related model, we conducted experiments showing that even with 0-cost or low-cost training (10\% parameter, 10\% data size) to obtain the task-related model, \textbf{ALPS} still achieves competitive performance, as shown in Table~\ref{tab:task_model_ablation}.

\begin{table*}[h!]
    \centering
    \begin{tabular}{cccccccc}
\toprule[1.2pt]
\multirow{2}{*}{\textbf{Method}} & \multirow{2}{*}{\textbf{$\mathcal{M}_\mathcal{T}$}} & \multicolumn{2}{c}{\textbf{Math}} & \multicolumn{4}{c}{\textbf{Code}} \\
\cmidrule(lr){3-4} \cmidrule(lr){5-8}
& & GSM8K & MATH & HEval & HEval+ & MBPP & MBPP+ \\
\midrule
\multicolumn{8}{c}{\textit{\textbf{Llama3.2-1B}}} \\
\midrule
w/full                     & SFT               & 17.82 & 4.60  & 45.12 & 39.08 & 28.63 & 23.78 \\
ALPS\_Readily      & Readily $\mathcal{M}$    & 21.81 & 6.87  & 45.78 & 39.29 & 27.93 & 24.01 \\
ALPS\_Fewer\_$\mathcal{D}$          & 10\% data & 22.96 & 7.06  & 46.91 & 39.87 & 28.73 & 23.62 \\
ALPS\_Fewer\_$\theta$      & attn only (10\%)  & \textbf{23.85} & 7.19  & \textbf{47.23} & \textbf{41.01} & 28.87 & 23.98 \\

\rowcolor{COLOR_MEAN}
\textbf{\textit{ALPS}}                       & SFT               & 23.74 & \textbf{7.28}  & 46.89 & 40.22 & \textbf{29.13} & \textbf{24.21} \\
\midrule
\multicolumn{8}{c}{\textit{\textbf{Llama3.2-3B}}} \\
\midrule
w/full                     & SFT               & 31.77 & 16.68 & 56.18 & 50.53 & 50.00 & 41.50 \\
ALPS\_Readily              & Readily $\mathcal{M}$    & 33.28 & 16.23 & 58.27 & 51.42 & \textbf{54.87} & \textbf{43.33} \\
ALPS\_Fewer\_$\mathcal{D}$ & 10\% data & 33.87 & 16.98 & 56.97 & 49.68 & 52.77 & 41.98 \\
ALPS\_Fewer\_$\theta$      & attn only (10\%)  & 34.17 & \textbf{17.33} & 56.89 & 50.05 & 51.79 & 40.32 \\

\rowcolor{COLOR_MEAN}
\textbf{\textit{ALPS}}                       & SFT               & \textbf{34.27} & 17.12 & \textbf{58.28} & \textbf{51.67} & 52.21 & 42.78 \\
\bottomrule[1.2pt]
\end{tabular}

    \caption{Experiments on variants of task-specific model acquisition.
    }
    \label{tab:task_model_ablation}
\end{table*}

These results illustrate that training a model via head identification using an existing task-related model (0-cost) yields comparable performance, even though these readily available models may not have been fully trained. Moreover, a task model obtained with less data or fewer parameters (lower cost) also performs well after \textbf{ALPS} head identification, demonstrating that ALPS does not necessarily require a complete full fine-tuning cycle to obtain the task-related model.

In summary, \textbf{ALPS is a method focused on efficient task head identification}. The task-related model serves merely as a component in the pipeline and \textbf{the acquisition process does not necessarily require a full fine-tuning cycle,} especially when leveraging readily available task models.

\section{Efficiency of \textbf{ALPS}}
To better understand the efficiency of \textbf{ALPS}, we list the training time cost in Table~\ref{tab:efficiency}. Our method shows significant reductions in training time, thus improving the efficiency.

\begin{table}[h!]
    \centering
    \resizebox{\linewidth}{!}{%
        \begin{tabular}{ccc}
\toprule[1.2pt]
\textbf{Model} & \textbf{Method} & \textbf{Avg.\ Time Cost (hrs)} \\ 
\midrule
\multirow[t]{2}{*}{Llama-3.2-1B} & w/ full & 2.02 \\
& \cellcolor{COLOR_MEAN}\textbf{\textit{ALPS}} & \cellcolor{COLOR_MEAN}\textbf{0.40} \\

\midrule
\multirow[t]{2}{*}{Llama-3.2-3B} & w/ full & 4.40 \\
& \cellcolor{COLOR_MEAN}\textbf{\textit{ALPS}} & \cellcolor{COLOR_MEAN}\textbf{1.05} \\

\midrule
\multirow[t]{2}{*}{Llama-3.1-8B} & w/ full & 7.67 \\
& \cellcolor{COLOR_MEAN}\textbf{\textit{ALPS}} & \cellcolor{COLOR_MEAN}\textbf{2.18} \\

\bottomrule[1.2pt]
\end{tabular}
    }
    \caption{Training time cost.
    }
    \label{tab:efficiency}
\end{table}

\section{Why is Parameter Alignment Distribution Score Better?}
Let us assume that the change from $\boldsymbol{W}^{h} _ {o,B}$ to $\boldsymbol{W}^{h} _ {o,T}$ can be modeled as a small translation in the parameter space: 
\begin{equation}
    \boldsymbol{W}^{h} _ {o,T} = \boldsymbol{W}^{h} _ {o,B} + \Delta,
\end{equation}
where $\Delta$ is a small perturbation that reflects task-specific adjustments. The W1 distance between $\boldsymbol{P} _ \mathcal{T}^{h}$ and $\boldsymbol{P} _ \mathcal{B}^{h}$ is defined as: 
\begin{equation}
    W_1\left( \boldsymbol{P} _ \mathcal{B}^{h}, \boldsymbol{P} _ \mathcal{T}^{h} \right)  =\inf_{\gamma \in \Gamma(\boldsymbol{P} _ \mathcal{B}^{h}, \boldsymbol{P} _ \mathcal{T}^{h})} \mathbb{E} _ {(x,y) \sim \gamma} \left[ \|x - y\| \right].
\end{equation}

For the special case where $\boldsymbol{P} _ \mathcal{T}^{h}$ is simply a translated version of $\boldsymbol{P} _ \mathcal{B}^{h}$, i.e., $\boldsymbol{P} _ \mathcal{T}^{h} \approx \boldsymbol{P} _ \mathcal{B}^{h}(x-\Delta)$, leading to:
\begin{equation}
    W_1\left( \boldsymbol{P} _ \mathcal{B}^{h}, \boldsymbol{P} _ \mathcal{T}^{h} \right) \approx \|\Delta \|.
\end{equation}

Thus, in the context of ALPS, $s _ {h}^{PAD} = W_1\left( \boldsymbol{P} _ \mathcal{B}^{h}, \boldsymbol{P} _ \mathcal{T}^{h} \right)$ directly reflects the magnitude of the parameter shift caused by task-specific fine-tuning. This linear relationship ensures that even small shifts in the attention head’s parameters are captured in proportion to $\|\Delta \|$. 

As for the KL divergence, the shifts between $\boldsymbol{P} _ \mathcal{T}^{h}$ and $\boldsymbol{P} _ \mathcal{B}^{h}$ is defined as: 
\begin{equation}
    D_{KL}(\boldsymbol{P} _ \mathcal{B}^{h} \| \boldsymbol{P} _ \mathcal{T}^{h}) = \sum _ {i} \boldsymbol{P} _ \mathcal{B}^{h}(i) \log \frac{\boldsymbol{P} _ \mathcal{B}^{h}(i)}{\boldsymbol{P} _ \mathcal{T}^{h}(i)}.
\end{equation}

For small perturbations $\Delta$, we can perform a second-order Taylor expansion. Under regularity conditions, this yields: 
\begin{equation}
    D_{KL}(\boldsymbol{P} _ \mathcal{B}^{h} \| \boldsymbol{P} _ \mathcal{T}^{h}) \approx \frac{1}{2}\,\mathcal{I}(\boldsymbol{P} _ \mathcal{B}^{h}) \, \|\Delta\|^2, 
\end{equation}
where $\mathcal{I}(\boldsymbol{P} _ \mathcal{B}^{h})$ is the Fisher information associated with $\boldsymbol{P} _ \mathcal{B}^{h}$.

Notice that while $W _ 1$ scales linearly with the magnitude of the shift $\| \Delta \|$, the KL divergence scales quadratically. This quadratic behavior means that for small but significant shifts, KL divergence may underemphasize these changes compared to $W _ 1$~\cite{arjovsky2017wassersteingan}. Moreover, KL divergence is asymmetric and can become unstable or even infinite if $\boldsymbol{P} _ \mathcal{T}^{h}$ assigns near-zero probability where $\boldsymbol{P} _ \mathcal{B}^{h}$ does not.

\section{Heatmaps}\label{appx_sec:heatmaps}
\begin{figure*}[h!]
    \centering
    \includegraphics[width=\linewidth]{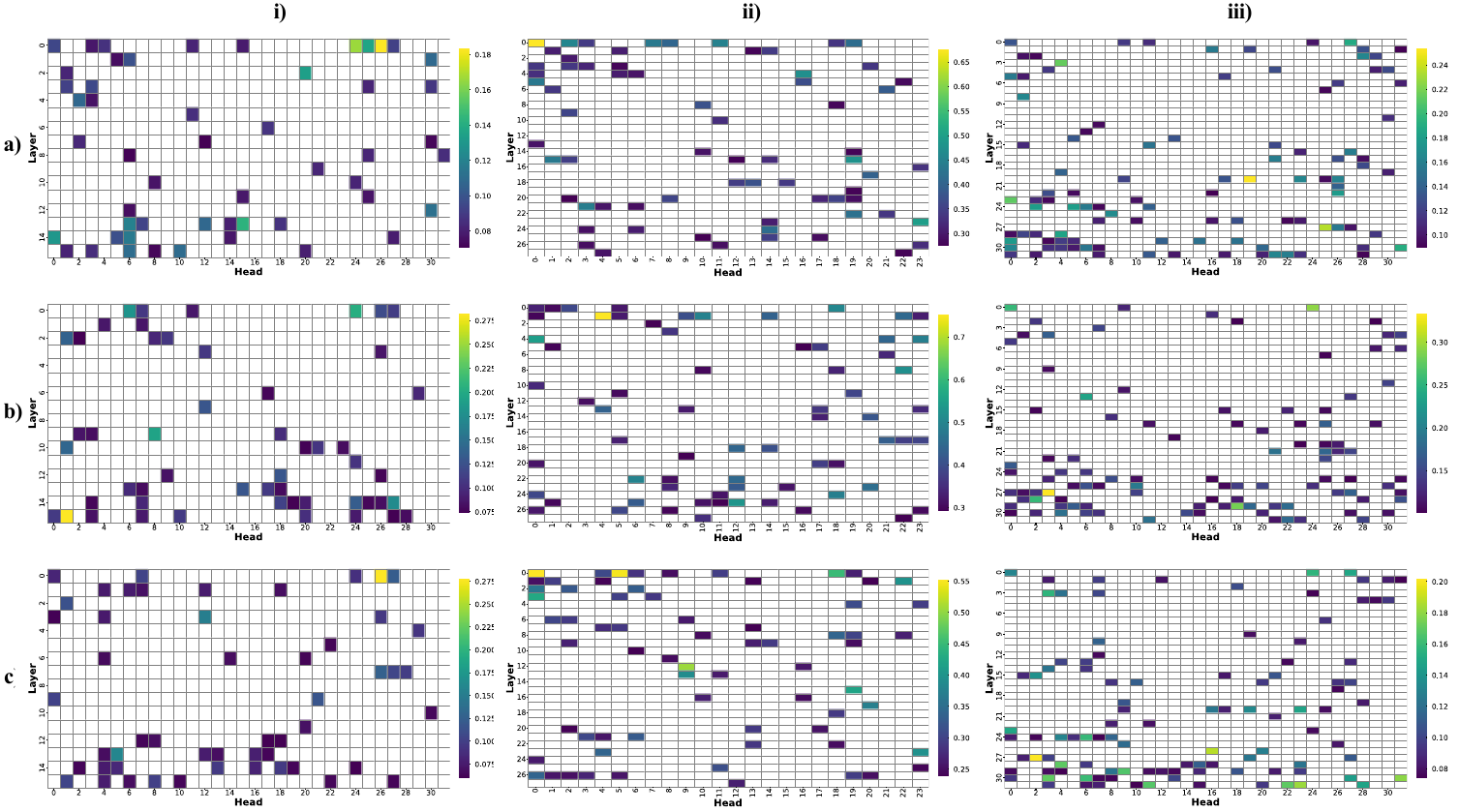}
    \caption{Heatmaps of task-sensitive attention heads (10\%) selected by our method on Llama-3.2-1B, Llama-3.2-3B, and Llama-3.1-8B, for \textbf{a)} general, \textbf{b)} math, and \textbf{c)} code tasks. The color intensity indicates the value of $s_{h}^{PAD}$ of each head.}
    \label{fig:full_heatmaps}
\end{figure*}

Figure~\ref{fig:full_heatmaps} presents heatmaps of the top 10\% task-sensitive attention heads selected by our method for three Llama-3 models (Llama-3.2-1B, Llama-3.2-3B, and Llama-3.1-8B) across general, math, and code tasks. Each cell in the heatmap corresponds to a specific layer-head combination, and its color intensity reflects the $s^{PAD}$ value. In the 1B and 3B models, the heatmaps for general and code tasks exhibit considerable overlap, whereas in the 8B model, the distributions for all three tasks are distinctly different. These observations indicate that as the model scale increases, the set of task-sensitive heads shifts, reflecting the diverse strategies that different models employ to handle task-specific requirements. Our method, ALPS, adaptively identifies these critical heads, demonstrating robust generalizability across various model scales and datasets.

\section{Full Tables}\label{sec:full_tables}
In this section, we present the complete baseline results corresponding to those summarized in Table~\ref{tab:head_transfer} and Table~\ref{tab:forgetting}.

\begin{table}[t]
    \centering
    \resizebox{\linewidth}{!}{%
        \begin{tabular}{cccccl}
\toprule[1.2pt]

\textbf{Model} & \textbf{Method} & \textbf{General} & \textbf{Math} & \textbf{Code} & \textbf{Avg.} \\

\midrule

\multirow{4}{*}{Llama-3.2-1B} 
& w/ full 
    & 14.28 & 5.87 & 22.28 & 14.14 \\
\cmidrule(lr){2-6}
& w/o attn 
    & 19.73 & 8.39 & 19.98 & 16.03 \footnotesize{\textcolor{teal}{$\uparrow$1.89}} \\
& Random 
    & 21.80 & 7.91 & 19.63 & 16.45 \footnotesize{\textcolor{teal}{$\uparrow$2.31}} \\
& LC
    & 21.57 & 8.02 & 22.23 & 17.27 \footnotesize{\textcolor{teal}{$\uparrow$3.13}} \\
& LoRA
    & 22.98 & 9.23 & 23.27 & 18.49 \footnotesize{\textcolor{teal}{$\uparrow$4.35}} \\
\rowcolor{COLOR_MEAN}
\cellcolor[HTML]{FFFFFF} 
& \textit{\textbf{ALPS}} 
    & \textbf{23.43} & \textbf{11.51} & \textbf{24.11} & \textbf{19.68 \footnotesize{\textcolor{teal}{$\uparrow$5.54}}} \\

\midrule

\multirow{4}{*}{Llama-3.2-3B} 
& w/ full 
    & 21.57 & 15.23 & 38.55 & 25.12 \\
\cmidrule(lr){2-6}
& w/o attn 
    & 22.31 & 13.49 & 35.72 & 23.84 \footnotesize{\textcolor{purple}{$\downarrow$1.28}} \\
& Random 
    & 22.38 & 17.80 & \textbf{41.37} & 27.18 \footnotesize{\textcolor{teal}{$\uparrow$2.06}} \\
& LC
    & 23.42 & 17.20 & 40.28 & 26.97 \footnotesize{\textcolor{teal}{$\uparrow$1.85}} \\
& LoRA
    & 24.01 & 17.98 & 40.57 & 27.52 \footnotesize{\textcolor{teal}{$\uparrow$2.40}} \\
\rowcolor{COLOR_MEAN}
\cellcolor[HTML]{FFFFFF} 
& \textit{\textbf{ALPS}} 
    & \textbf{24.65} & \textbf{18.70} & 40.24 & \textbf{27.86 \footnotesize{\textcolor{teal}{$\uparrow$2.74}}} \\

\midrule

\multirow{4}{*}{Llama-3.1-8B} 
& w/ full 
    & 27.21 & 28.76 & 42.34 & 32.77 \\
\cmidrule(lr){2-6}
& w/o attn 
    & 18.29 & 25.77 & \textbf{46.11} & 30.06 \footnotesize{\textcolor{purple}{$\downarrow$3.38}} \\
& Random 
    & 27.93 & \textbf{32.96} & 40.43 & 33.77 \footnotesize{\textcolor{teal}{$\uparrow$1.00}} \\
& LC
    & 26.52 & 30.28 & 41.28 & 32.69 \footnotesize{\textcolor{purple}{$\downarrow$0.08}} \\
& LoRA
    & 27.87 & 31.02 & 42.27 & 33.72 \footnotesize{\textcolor{teal}{$\uparrow$0.95}} \\
\rowcolor{COLOR_MEAN}
\cellcolor[HTML]{FFFFFF} 
& \textit{\textbf{ALPS}} 
    & \textbf{28.21} & 31.31 & 44.57 & \textbf{34.67 \footnotesize{\textcolor{teal}{$\uparrow$1.90}}} \\

\bottomrule[1.2pt]

\end{tabular}
    }
    \caption{Full table of Table~\ref{tab:head_transfer}.
    }
    \label{tab:full_head_transfer}
\end{table}

\begin{table}[t]
    \centering
    \resizebox{0.8\linewidth}{!}{%
        \begin{tabular}{ccccc}
\toprule[1.2pt]

\textbf{Model} & \textbf{Method} & \textbf{MMLU} & \textbf{ARC-C} & \textbf{Avg.} \\

\midrule

\multirow{5}{*}{Llama-3.2-1B} 
& vanilla 
    & 32.2 & 32.8 & 32.5 \\
\cmidrule(lr){2-5}
& w/ full 
    & 28.14 & 26.95 & 27.55  \\
& w/o attn 
    & 27.80 & 23.18 & 25.49 \\
& Random 
    & 27.86 & 26.27 & 27.07 \\
& LC 
    & 26.83 & 27.08 & 26.96 \\
& LoRA
    & 28.79 & 27.98 & 28.39 \\
\rowcolor{COLOR_MEAN}
\cellcolor[HTML]{FFFFFF} 
& \textit{\textbf{ALPS}} 
    & \textbf{29.88} & \textbf{28.73} & \textbf{29.31} \\

\midrule

\multirow{5}{*}{Llama-3.2-3B} 
& vanilla 
    & 58 & 69.1 & 63.55 \\
\cmidrule(lr){2-5}
& w/ full 
    & 44.22 & 50.82 & 47.52 \\
& w/o attn 
    & 46.16 & 46.95 & 46.56 \\
& Random 
    & 45.55 & 47.30 & 46.43 \\
& LC 
    & 44.34 & 48.98 & 46.66 \\
& LoRA
    & 47.28 & 50.88 & 49.08 \\
\rowcolor{COLOR_MEAN}
\cellcolor[HTML]{FFFFFF} 
& \textit{\textbf{ALPS}} 
    & \textbf{48.83} & \textbf{52.37} & \textbf{50.60} \\

\midrule

\multirow{5}{*}{Llama-3.1-8B} 
& vanilla 
    & 66.7 & 79.7 & 73.2 \\
\cmidrule(lr){2-5}
& w/ full 
    & 55.40 & 60.34 & 57.87  \\
& w/o attn 
    & 22.95 & 49.01 & 35.98  \\
& Random 
    & 54.59 & 62.92 & 58.76  \\
& LC 
    & 53.82 & 61.89 & 57.86 \\
& LoRA
    & 56.28 & 63.01 & 59.65 \\
\rowcolor{COLOR_MEAN}
\cellcolor[HTML]{FFFFFF} & 
\textit{\textbf{ALPS}} 
    & \textbf{57.87} & \textbf{64.29} & \textbf{61.08}  \\

\bottomrule[1.2pt]

\end{tabular}
    }
    \caption{Full table of Table~\ref{tab:forgetting}.}
    \label{tab:full_forgetting}
\end{table}

In Table~\ref{tab:main_table} we mentioned that each task employs a separately fine-tuned model trained on its own dataset. With six baselines, three datasets, and three model scales, this setup requires $6\times3\times3 = 54$ distinct models, making a full evaluation across ten benchmarks computationally prohibitive. Consequently, our main experiments report only per-task performance. To probe ALPS’s generalization further, we conducted selective evaluations of the baselines on additional tasks. Table~\ref{tab:full_main_table} summarizes these results. These findings confirm that ALPS consistently generalizes better across diverse tasks.

\begin{table*}[t]
    \centering
    \resizebox{\linewidth}{!}{%
        \begin{tabular}{cccccccccccl}
\toprule[1.2pt]
\multirow{2}{*}{\textbf{Model}} & \multirow{2}{*}{\textbf{Method}} & \multirow{2}{*}{\textbf{Task}} & \multirow{2}{*}{\textbf{Attn\%}} & \multicolumn{2}{c}{\textbf{General}} & \multicolumn{2}{c}{\textbf{Math}} & \multicolumn{4}{c}{\textbf{Code}} \\
\cmidrule(lr){5-6} \cmidrule(lr){7-8} \cmidrule(lr){9-12}
& & & & IFEval & GPQA & GSM8K & MATH & HEval & HEval+ & MBPP & MBPP+ \\

\midrule
\multirow{3}{*}{1B-MC} 
  & w/ full    & code & 100\% 
    & 27.58 & 10.27 & \textbf{7.88}  & \textbf{2.50}  
    & \textcolor{violet}{\textit{45.12}} & \textcolor{violet}{\textit{39.08}} & \textcolor{violet}{\textit{28.63}} & \textcolor{violet}{\textit{23.78}} \\
  & w/o attn   & code &   0\% 
    & 28.30 & 11.16 & 7.28           & 2.10           
    & \textcolor{violet}{\textit{41.51}} & \textcolor{violet}{\textit{35.28}} & \textcolor{violet}{\textit{\textbf{29.91}}} & \textcolor{violet}{\textit{\textbf{24.88}}} \\
\rowcolor{COLOR_MEAN}\cellcolor[HTML]{FFFFFF}
  & \textit{\textbf{ALPS}} & code & 10\% 
    & \textbf{29.38} & \textbf{12.27} & 7.81   & 2.48   
    & \textcolor{violet}{\textit{\textbf{46.89}}} & \textcolor{violet}{\textit{\textbf{40.22}}} & \textcolor{violet}{\textit{29.13}} & \textcolor{violet}{\textit{24.21}} \\
\midrule

\multirow{3}{*}{1B-MI} 
  & w/ full    & math & 100\% 
    & 31.06 &  8.26 & \textcolor{violet}{\textit{17.82}} &  \textcolor{violet}{\textit{4.60}}  
    & 11.63 & 10.42 & 16.91 & \textbf{16.17} \\
  & w/o attn   & math &   0\% 
    & 30.18 & 17.63 & \textcolor{violet}{\textit{23.35}} &  \textcolor{violet}{\textit{5.20}}  
    & 12.87 & 11.67 & 16.92 & 14.87 \\
\rowcolor{COLOR_MEAN}\cellcolor[HTML]{FFFFFF}
  & \textit{\textbf{ALPS}} & math & 10\% 
    & \textbf{32.82} & \textbf{18.72} & \textcolor{violet}{\textit{\textbf{23.74}}} & \textcolor{violet}{\textit{\textbf{7.28}}}  
    & \textbf{16.55} & \textbf{14.07} & \textbf{19.38} & 15.68 \\
\midrule

\multirow{3}{*}{1B-UC} 
  & w/ full    & general & 100\% 
    & \textcolor{violet}{\textit{34.53}} & \textcolor{violet}{\textit{18.75}} & \textbf{6.93} & 1.24  
    & 17.17 & 14.62 & 18.89 & 14.83 \\
  & w/o attn   & general &   0\% 
    & \textcolor{violet}{\textit{35.85}} & \textcolor{violet}{\textit{19.64}} & 5.84          & 2.52  
    & \textbf{17.77} & \textbf{15.21} & 17.75 & 12.72 \\
\rowcolor{COLOR_MEAN}\cellcolor[HTML]{FFFFFF}
  & \textit{\textbf{ALPS}} & general & 10\% 
    & \textcolor{violet}{\textit{\textbf{36.69}}} & \textcolor{violet}{\textit{\textbf{26.16}}} & 6.44 & \textbf{3.32}  
    & 16.55 & 14.38 & \textbf{19.32} & \textbf{15.67} \\
\midrule

\multirow{3}{*}{3B-MC} 
  & w/ full    & code & 100\% 
    & \textbf{40.41} & \textbf{18.32} 
    & 28.73 & 8.72  
    & \textcolor{violet}{\textit{56.18}} & \textcolor{violet}{\textit{50.53}} & \textcolor{violet}{\textit{50.00}} & \textcolor{violet}{\textit{41.50}} \\
  & w/o attn   & code &   0\% 
    & 37.65 & 17.41 
    & 31.31 & 9.44  
    & \textcolor{violet}{\textit{55.52}} & \textcolor{violet}{\textit{50.62}} & \textcolor{violet}{\textit{48.71}} & \textcolor{violet}{\textit{39.72}} \\
\rowcolor{COLOR_MEAN}\cellcolor[HTML]{FFFFFF}
  & \textit{\textbf{ALPS}} & code & 10\% 
    & 37.65 & 18.08 
    & \textbf{32.26} & \textbf{10.28}  
    & \textcolor{violet}{\textit{\textbf{58.28}}} & \textcolor{violet}{\textit{\textbf{51.67}}} & \textcolor{violet}{\textit{\textbf{52.21}}} & \textcolor{violet}{\textit{\textbf{42.78}}} \\
\midrule

\multirow{3}{*}{3B-MI} 
  & w/ full    & math & 100\% 
    & 35.25 & 17.63 
    & \textcolor{violet}{\textit{31.77}} & \textcolor{violet}{\textit{16.68}}  
    & 26.28 & 23.21 & 29.92 & 25.73 \\
  & w/o attn   & math &   0\% 
    & \textbf{35.37} & 17.41 
    & \textcolor{violet}{\textit{33.21}} & \textcolor{violet}{\textit{13.76}}  
    & 28.27 & 25.63 & 31.72 & 27.79 \\
\rowcolor{COLOR_MEAN}\cellcolor[HTML]{FFFFFF}
  & \textit{\textbf{ALPS}} & math & 10\% 
    & 33.33 & \textbf{18.33} 
    & \textcolor{violet}{\textit{\textbf{34.27}}} & \textcolor{violet}{\textit{\textbf{17.12}}}  
    & \textbf{29.93} & \textbf{26.82} & \textbf{34.11} & \textbf{28.82} \\
\midrule

\multirow{3}{*}{3B-UC} 
  & w/ full    & general & 100\% 
    & \textcolor{violet}{\textit{42.81}} & \textcolor{violet}{\textit{22.32}} 
    & 6.82 & 8.16  
    & 29.92 & 27.43 & 33.33 & 29.13 \\
  & w/o attn   & general &   0\% 
    & \textcolor{violet}{\textit{40.29}} & \textcolor{violet}{\textit{\textbf{24.33}}} 
    & 6.62 & \textbf{8.22}  
    & \textbf{32.94} & \textbf{28.49} & \textbf{34.97} & 28.82 \\
\rowcolor{COLOR_MEAN}\cellcolor[HTML]{FFFFFF}
  & \textit{\textbf{ALPS}} & general & 10\% 
    & \textcolor{violet}{\textit{\textbf{44.96}}} & \textcolor{violet}{\textit{\textbf{24.33}}} 
    & 5.98 & 7.92  
    & 29.33 & 26.82 & 34.43 & \textbf{29.97} \\
\midrule

\multirow{3}{*}{8B-MC} 
  & w/ full    & code & 100\% 
    & 47.84 & 18.75 
    & 46.63 & 13.40  
    & \textcolor{violet}{\textit{69.50}} & \textcolor{violet}{\textit{64.28}} & \textcolor{violet}{\textit{63.21}} & \textcolor{violet}{\textit{52.38}} \\
  & w/o attn   & code &   0\% 
    & 47.48 & 19.20 
    & \textbf{49.13} & 10.22  
    & \textcolor{violet}{\textit{70.72}} & \textcolor{violet}{\textit{\textbf{65.88}}} & \textcolor{violet}{\textit{63.27}} & \textcolor{violet}{\textit{52.55}} \\
\rowcolor{COLOR_MEAN}\cellcolor[HTML]{FFFFFF}
  & \textit{\textbf{ALPS}} & code & 10\% 
    & \textbf{49.96} & \textbf{19.64} 
    & 48.98 & \textbf{15.48}  
    & \textcolor{violet}{\textit{\textbf{72.68}}} & \textcolor{violet}{\textit{65.03}} & \textcolor{violet}{\textit{\textbf{63.67}}} & \textcolor{violet}{\textit{\textbf{52.88}}} \\
\midrule

\multirow{3}{*}{8B-MI} 
  & w/ full    & math & 100\% 
    & 36.69 & 20.76 
    & \textcolor{violet}{\textit{63.00}} & \textcolor{violet}{\textit{20.52}}  
    & 30.53 & 28.78 & 28.23 & 23.80 \\
  & w/o attn   & math &   0\% 
    & 26.38 & 22.78 
    & \textcolor{violet}{\textit{64.06}} & \textcolor{violet}{\textit{15.48}}  
    & 32.38 & 31.28 & 30.13 & 27.62 \\
\rowcolor{COLOR_MEAN}\cellcolor[HTML]{FFFFFF}
  & \textit{\textbf{ALPS}} & math & 10\% 
    & \textbf{38.13} & \textbf{24.11} 
    & \textcolor{violet}{\textit{\textbf{64.13}}} & \textcolor{violet}{\textit{\textbf{22.48}}}  
    & \textbf{35.43} & \textbf{32.33} & \textbf{34.18} & \textbf{28.37} \\
\midrule

\multirow{3}{*}{8B-UC} 
  & w/ full    & general & 100\% 
    & \textcolor{violet}{\textit{51.20}} & \textcolor{violet}{\textit{25.22}} & 54.06 & 15.12  
    & 32.33 & 28.72 & 41.54 & 34.92 \\
  & w/o attn   & general &   0\% 
    & \textcolor{violet}{\textit{40.41}} & \textcolor{violet}{\textit{26.16}} & 53.28 & 14.28  
    & 33.28 & 28.62 & 41.58 & 34.78 \\
\rowcolor{COLOR_MEAN}\cellcolor[HTML]{FFFFFF}
  & \textit{\textbf{ALPS}} & general & 10\% 
    & \textcolor{violet}{\textit{\textbf{51.33}}} & \textcolor{violet}{\textit{\textbf{27.29}}} & \textbf{54.28} & \textbf{15.48}  
    & \textbf{35.42} & \textbf{29.37} & \textbf{42.33} & \textbf{36.21} \\

\bottomrule[1.2pt]
\end{tabular}
    }
    \caption{Selective baseline evaluation across additional tasks. Italicized and correspond to those in Table~\ref{tab:main_table}
    }
    \label{tab:full_main_table}
\end{table*}

\end{document}